\documentclass[preprint,review,12pt]{elsarticle}

\usepackage{amssymb}
\usepackage{amsmath}

\usepackage{booktabs}
\usepackage{multirow}
\usepackage{siunitx}
\usepackage{float}

\journal{Pattern Recognition}

\begin{document}

\begin{frontmatter}



\title{Adaptive receptive field-based spatial-frequency feature reconstruction network for fine-grained few-shot image classification}




\author[1]{Linyue Zhang}
\author[1]{Wenyi Zeng}
\author[2]{Zicheng Pan}
\author[3]{Yongsheng Gao}
\author[4]{Changming Sun}
\author[5]{Jun Hu}
\author[6]{Lixian Liu}
\author[1]{Weichuan Zhang\corref{cor1}}
\author[7]{Tuo Wang\corref{cor1}}

\cortext[cor1]{Corresponding author.}

%
%

\address[1]{School of Electronic Information and Artificial Intelligence, Shaanxi University of Science and Technology, Xi'an, Shaanxi Province, China}
\address[2]{School of Electrical Engineering and Computer Science, the University of Queensland, QLD, Australia}
\address[3]{Institute for Integrated and Intelligent Systems, Griffith University, QLD, Australia}
\address[4]{CSIRO Data61, PO Box 76, Epping, NSW 1710, Australia}
\address[5]{Chengdu University of Technology, Chengdu, Sichuan Province, China}
\address[6]{Xidian University, Xi'an, Shaanxi Province, China}
\address[7]{The First Affiliated Hospital of Xi'an Jiaotong University, Xi'an, Shaanxi Province, China}

\begin{abstract}
Feature reconstruction techniques are widely applied for few-shot fine-grained image classification (FSFGIC). Our research indicates that one of the main challenges facing existing feature-based FSFGIC methods is how to choose the size of the receptive field to extract feature descriptors (including spatial and frequency feature descriptors) from different category input images, thereby better performing the FSFGIC tasks. To address this, an adaptive receptive field-based spatial-frequency feature reconstruction network (ARF-SFR-Net) is proposed. The designed ARF-SFR-Net has the capability to adaptively determine receptive field sizes for obtaining spatial and frequency features, and effectively fuse them for reconstruction and FSFGIC tasks. The designed ARF-SFR-Net can be easily embedded into a given episodic training mechanism for end-to-end training from scratch. Extensive experiments on multiple FSFGIC benchmarks demonstrate the effectiveness and superiority of the proposed ARF-SFR-Net over state-of-the-art approaches. The code is available at: https://github.com/ICL-SUST/ARF-SFR-Net.git.

\end{abstract}




\end{frontmatter}



\section{Introduction}
\label{sec1}
Fine-grained image classification~\cite{liao2026neuron, liao2025dynamic,  jing2022recent} aims to distinguish visually similar object subcategories, such as birds~\cite{wah2011caltech}, dogs~\cite{khosla2011novel}, or cars~\cite{krause20133d} on different datasets. Although convolutional neural networks (CNNs) achieve remarkable performance across various tasks~\cite{zhang2023image,ma2023ct,liu2024aekan,li2019multi, islam2023background, li2023mutual, wang2018survey, an2023edge, li2023m, wang2020corner, zhang2023image, qiu2021recurrent, li2023traffic, zhang2019corner, zhang2014corner, zhang2019discrete, zhang2020corner, zhang2015contour, zhang2017noise, shui2012noise, 6507646, guo2026gattenrnn, song2025efficient, liao2025visual, huang2022heart, tang2025cascading, lu2026meningioma}, their success heavily depends on large annotated datasets~\cite{ren2025adaptive, lu2023track}. However, annotations in few-shot fine-grained domains are scarce, expensive, and predominantly manual. Consequently, few-shot fine-grained image classification (FSFGIC)~\cite{zhang2024re, wang2024unbiased, lei2024semi, jing2021novel, liao2025dynamic, wang2025principal, liao2022asrsnet, lu2022image, zhang2021ndpnet, ren2024few, pan2024pseudo, ren2025zero,zhang2026simultaneous, wang2026dual} has emerged as an important yet highly challenging task that demands efficient feature learning from only a few samples per class.
\begin{figure}[h]
	\centering
	\includegraphics[width=0.6\linewidth]{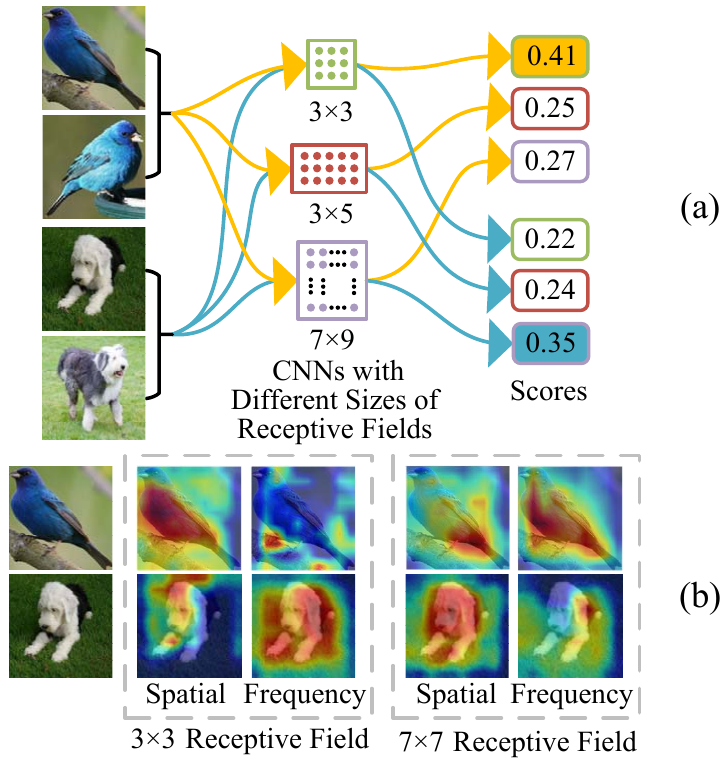}
	\caption{(a) The impact of receptive field size on the FSFGIC accuracy of BDFRNet~\cite{wu2024bi1}, and (b) The impact of receptive field size on the spatial- and frequency-domain feature representations of BDFRNet~\cite{wu2024bi1}.}
	\label{fig:1}
\end{figure}

Feature reconstruction techniques~\cite{Wertheimer2021FRN} are widely used in FSFGIC for effectively analyzing image feature representations by exploring the relationship between different images. Our research indicates that one of the main challenges facing existing feature reconstruction-based FSFGIC methods is how to choose the size of the receptive field to extract feature descriptors (including spatial and frequency feature descriptors) from different category input images, thereby better performing the FSFGIC tasks. Using examples of birds~\cite{wah2011caltech} and dogs~\cite{khosla2011novel} (two images per category/breed) as shown in Fig.~\ref{fig:1}, BDFRNet~\cite{wu2024bi1} is used as the backbone. From Fig.~\ref{fig:1}(a), it is observed that varying receptive field sizes significantly influence classification accuracy. The highest inter-bird similarity and lowest inter-dog similarity occur with a receptive field of size 3$\times$3. Conversely, a receptive field of size 7$\times$7 yields the lowest inter-bird but the highest inter-dog similarity. Furthermore, the gradient-weighted class activation mapping (Grad-CAM) visualization technique~\cite{selvaraju2017grad, zheng2023fully, jing2023ecfrnet} is employed for illustrating the domain-specific feature quality under different receptive field sizes, showing that the spatial and frequency branches attend to different cues in a receptive-field-dependent manner, as shown in Fig.~\ref{fig:1}(b). With a 3$\times$3 receptive field, spatial features cleanly highlight the bird target but are noisy for the dog, while frequency features excel at focusing on the dog but are susceptible to background noise for the bird. Conversely, a 7$\times$7 receptive field reverses this effect: spatial features now precisely capture the dog, whereas frequency features effectively highlight the bird.

To address the aforementioned problems, we propose an adaptive receptive field-based spatial–frequency feature reconstruction framework for FSFGIC. Specifically, an adaptive receptive field (ARF) strategy is developed to dynamically predict and adjust the geometric structure of receptive fields in both spatial and frequency domains for achieving high-quality feature representations from images. Then a feature fusion scheme is designed for integrating spatial- and frequency-domain descriptors for reliable feature reconstruction. Based on this design, a novel adaptive receptive field-based spatial-frequency feature reconstruction network (ARF-SFR-Net) is introduced. The designed ARF-SFR-Net can be easily embedded into a given episodic training mechanism for end-to-end training from scratch. The proposed ARF-SFR-Net is compared with twenty-nine state-of-the-art methods. Five fine-grained image classification benchmark datasets (i.e., meta-iNat~\cite{wertheimer2019few}, tiered meta-iNat~\cite{wertheimer2019few}, CUB-200-2011~\cite{wah2011caltech}, Stanford Dogs~\cite{khosla2011novel}, and Stanford Cars~\cite{krause20133d}) are utilized to evaluate the performance of the proposed method in comparison with the benchmarks. The experimental results demonstrate the effectiveness and superiority of the proposed ARF-SFR-Net over twenty-nine state-of-the-art approaches on both 5-way 1-shot and 5-way 5-shot FSFGIC tasks. 

The rest of this paper is organized as follows. In Section 2, the background and related works are briefly introduced. A new adaptive receptive field-based spatial–frequency feature extraction (ARF-SFE) module and a new adaptive receptive field-based spatial-frequency feature reconstruction network (ARF-SFR-Net) are presented in Section 3. Experimental results are shown and analyzed in Section 4. Finally, conclusions are drawn in Section 5.

\section{Related Work}
\label{sec1}

\subsection{Optimization-based FSFGIC Methods}
\label{subsec1}

Based on the `learning to learn' paradigm, optimization-based FSFGIC methods quickly learn model initializations or parameter updates using three key techniques: attention mechanisms~\cite{ruan2021few, wang2024bi}, feature alignment~\cite{leng2024meta}, and knowledge distillation~\cite{wu2023few}. Attention-based methods learn distinctive descriptors through mechanisms such as 
spatial~\cite{ruan2021few} or dual-attention~\cite{Shulin2022} networks, transformer-based~\cite{zhang2022crossimage} architectures, and dual-channel attention~\cite{wang2024bi} to improve feature discrimination. Feature alignment methods spatially align objects between support and query images to capture fine-grained differences, using techniques like 
distribution alignment~\cite{leng2024meta}. Knowledge distillation enhances efficiency and generalization by transferring knowledge from complex to simpler models, often via meta-distillation frameworks~\cite{wu2023few}.

\subsection{Metric-learning Based FSFGIC Methods}
\label{subsec2}

Metric-based few-shot learning methods learn embedding functions that map samples into a feature space where classification is determined by a predefined or learned similarity measure~\cite{ren2025adaptive}.~Significant research focuses on sophisticated feature matching between query and support images. For example, Zhang et al.~\cite{zhang2022deepemd} developed DeepEMD, which formulates image matching as an optimal transport problem using Earth mover's distance. In a different approach, Wertheimer et al.~\cite{Wertheimer2021FRN} introduced a feature map reconstruction network (FRN) that reconstructs query features from support features via ridge regression. This was later extended by Wu et al.~\cite{wu2024bi1} to a bi-directional feature reconstruction network (BDFRNet) to increase inter-class variation and reduce intra-class variation. Furthermore, Ma et al.~\cite{ma2024cross} designed a cross-layer and cross-sample feature optimization network (C2-Net) to extract discriminative features across layers and address feature mismatches from both channel and spatial dimensions. Subsequently, Ma et al.~\cite{ma2025few} introduced SUITED, a method for progressive feature refinement and continuous relationship modeling to dynamically capture prototype-query associations. 
Attention mechanisms have been integrated into metric-based methods to enhance the discrimination of localized regions. 
Yang et al.~\cite{yang2024channel} proposed a channel-spatial cross-attention module to leverage inter-image spatial and channel information. 
Lee et al.~\cite{lee2024task} introduced a dual attention framework, featuring a support attention module to emphasize class-discriminative channels and a query attention module to weight object-relevant channels. 
Long et al.~\cite{long2026atars} proposed ATARS, which improves discriminative representation learning for few-shot fine-grained classification through adaptive task-aware feature learning.

\subsection{Spatial-frequency Representation and Fusion Methods}
\label{subsec3}

Recent studies demonstrate that leveraging complementary cues from the spatial and frequency domains can enhance fine-grained discriminability in few-shot learning, thereby supporting more reliable feature reconstruction and matching. For instance, FGFL~\cite{Cheng2023FGFL} systematically examines frequency band roles and employs frequency-guided training, while MEFP~\cite{Zhou2024MEFP} decomposes images into low- and high-frequency branches and aligns them with spatial features via reconstruction. FSC~\cite{Ji2025FSC} further achieves spatial-frequency complementarity via channel-wise style adaptation. In hyperspectral domains, methods such as SSSCD~\cite{Cao2024SSSCD, bao2022corner} jointly align spatial-spectral features, and TEFSL~\cite{Xi2025TEFSL} utilizes query statistics for transductive spectral–spatial embedding. 

Within the scope of our investigation, the existing feature reconstruction-based methods suffer from a key gap: no adaptive receptive field strategy addresses domain discrepancies and differing preferences between spatial and frequency cues. To bridge this, we propose ARF-SFR-Net, a network with a novel adaptive receptive field mechanism that independently learns domain-specific sampling for spatial and frequency branches. This extracts high-quality descriptors and enables adaptive fusion for feature reconstruction. By adapting receptive fields at the domain, scale, and instance levels, ARF-SFR-Net significantly boosts reconstruction reliability, fostering robust and discriminative representations under few-shot fine-grained constraints.

\section{Methodology}
\label{sec3}

\subsection{Problem Definition}
\label{subsec4}

Formally, given a dataset $\mathcal{D}$ comprising $\mathcal{L}$ fine-grained categories, it is partitioned into three disjoint subsets: a training set $\mathcal{D}_{\text{train}}$, a validation set $\mathcal{D}_{\text{val}}$, and a test set $\mathcal{D}_{\text{test}}$, such that ${D}_{\text{train}}$$\cap$${D}_{\text{val}}$$\cap$${D}_{\text{test}}$=$\emptyset$, $\mathcal{D}_{\text{train}}$$\cup$$\mathcal{D}_{\text{val}}$$\cup$$\mathcal{D}_{\text{test}}$=${\mathcal{D}}$. Here, $\mathcal{D}_{\text{train}}$ is used for training, while $\mathcal{D}_{\text{test}}$ consists of unseen fine-grained categories not encountered during training. In each FSFGIC episode, a support set $\mathcal{S}$ and a query set $\mathcal{Q}$ are constructed. The support set includes $N$ distinct classes, each with $K$ labeled samples. The query set contains unlabeled samples sharing the same label space as the support set. The primary goal is to learn an embedding function $f_{\theta}(\cdot)$ that maps an input image $X$ to a discriminative representation $z=f_{\theta}(X)$, such that each query sample can be accurately classified into its corresponding category in the support set. This framework defines a standard $\mathcal{N}$-way $K$-shot fine-grained classification task.

\subsection{Overall Framework}
\label{subsec5}
\begin{figure*}[htbp]
	\centering
	\includegraphics[width=1.\linewidth]{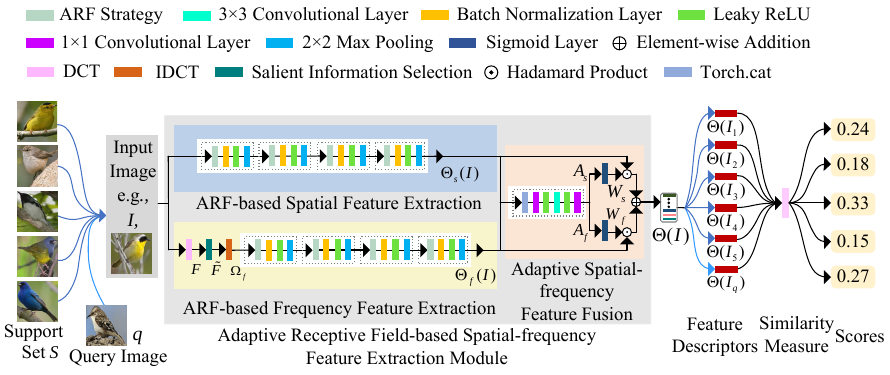}
	\caption{The pipeline of the proposed ARF-SFR-Net for a 5-way 1-shot FSFGIC task based on the Conv-4 backbone.}
	\label{fig:22}
\end{figure*}

As shown in Fig.~\ref{fig:22}, the proposed ARF-SFR-Net comprises two main components: An adaptive receptive field-based spatial–frequency feature extraction (ARF-SFE) module and a similarity measure module. Following mainstream FSFGIC frameworks such as DeepEMD~\cite{zhang2022deepemd}, FRN~\cite{Wertheimer2021FRN}, and SUITED~\cite{ma2025few}, the network employs Conv-4~\cite{snell20177} and ResNet-12~\cite{lee2019meta} as backbones. The ARF-SFE module incorporates a novel adaptive receptive field strategy into these backbones, enabling dynamic adjustment of receptive field size and shape across both spatial and frequency domains. The complementary features from the two branches are fused via pixel-wise weighted aggregation, which adaptively emphasizes the discriminative domain information to form robust feature representations. Finally, the similarity measure module computes distances between support and query samples to perform FSFGIC tasks.


\subsection{Conv-4-based ARF-SFE Module}
The designed ARF-SFE module includes a new adaptive receptive field (ARF) strategy, an ARF-based spatial-frequency feature extraction, and an adaptive spatial-frequency feature fusion, as illustrated below.

\subsubsection{A New Adaptive Receptive Field (ARF) Strategy}
\begin{figure*}[htbp]
	\centering
	\includegraphics[width=.9\linewidth]{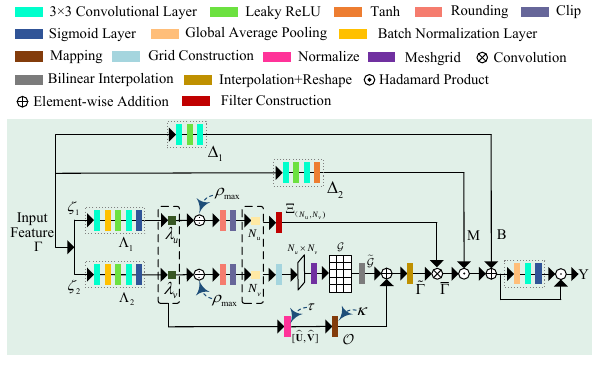}
	\caption{The proposed adaptive receptive field (ARF) strategy.}
	\label{fig:2}
\end{figure*}

According to the input images' content, the designed ARF strategy (as shown in Fig.~\ref{fig:2}) has the capability to adaptively adjust the kernel shape and receptive field size, enabling the designed network to effectively capture task-relevant structures even with extremely limited training samples.

Given an input feature map $\Gamma\in\mathbb{R}^{C\times H\times W}$ obtained from either the spatial or frequency branch, global average pooling (GAP) is applied to $\Gamma$ along the
vertical and horizontal axes for obtaining vertical and horizontal axes' global descriptors 
\begin{equation}
	\begin{aligned}
		&\zeta_1 = \mathrm{GAP}_{\text{vertical}}(\Gamma)\in \mathbb{R}^{C},\\
		&\zeta_2 = \mathrm{GAP}_{\text{horizontal}}(\Gamma)\in \mathbb{R}^{C}.
	\end{aligned}
\end{equation}

Subsequently, two lightweight scale prediction mappings (i.e., $\Lambda_1$ and $\Lambda_2$) are employed for estimating the vertical kernel scale $\lambda_u$ and the horizontal kernel scale $\lambda_v$
\begin{equation}
	\begin{aligned}
		\label{eq11}
		&\lambda_u = \mathrm{Sigmoid}\!\big(\mathrm{Conv}_{3\times3}(\ell(\beta(\mathrm{Conv}_{3\times3}(\zeta_1))))\big)\,(\rho_{\max}-1)+1,\\
		&\lambda_v = \mathrm{Sigmoid}\!\big(\mathrm{Conv}_{3\times3}(\ell(\beta(\mathrm{Conv}_{3\times3}(\zeta_2))))\big)\,(\rho_{\max}-1)+1,
	\end{aligned}
\end{equation}
where $\ell(\cdot)$ denotes the LeakyReLU activation, $\beta(\cdot)$ denotes the batch normalization operation, and $\rho_{\max}$ is the maximum selectable size of the kernel.

The height $N_u$ and width $N_v$ of the discrete receptive field are calculated as 
\begin{equation}
	\label{eq5}
	\begin{aligned}
		N_u = 2\Big\lfloor \frac{\lambda_u}{\varsigma}\Big\rfloor + 1,~N_v = 2\Big\lfloor \frac{\lambda_v}{\varsigma}\Big\rfloor + 1,
	\end{aligned}
\end{equation}
where $\varsigma$ is a step $\varsigma$$\in$$[1,3,..,\rho_{\max}]$ and $\lfloor\cdot\rfloor$ is the round down operation. 

The discrete kernel size $(N_u, N_v)$ has a dual role: it sets the resolution of the base sampling grid and selects the matching anisotropic convolution branch from a predefined kernel bank. We therefore first build a normalized sampling grid from $(N_u, N_v)$ and then apply the chosen convolution branch to the adaptively resampled features.

A normalized sampling grid with resolution $(N_u,N_v)$ is defined as
\begin{align}
	\mathcal{G}=\{(\xi_i,\eta_j)\mid \xi_i,\eta_j\in[-1,1]\},
\end{align}
where $\xi_i$ $(i=1,...,{N_u})$ and $\eta_j$ $(j=1,...,{N_v})$ denote the one-dimensional sampling coordinate sequences along the vertical and horizontal axes respectively, with each coordinate sampled uniformly over the interval $[-1,1]$. Bilinear interpolation operation is performed on $\mathcal{G}$ to make the feature resolution consistent with the input image 
\begin{equation}
	\begin{aligned}
		\label{eq130}
		\tilde{\mathcal{G}}=\mathrm{Upsample}(\mathcal{G};\,H,W)\in \mathbb{R}^{2\times H\times W}.\\
	\end{aligned}
\end{equation}

The proposed ARF strategy further converts the two scale fields into a continuous offset field for deforming the sampling grid as follows:
\begin{equation}
	\begin{aligned}
		\label{eq10}
		&\widehat{\mathbf{U}} = \frac{\lambda_u-\tau}{\tau},~\widehat{\mathbf{V}} = \frac{\lambda_v-\tau}{\tau}, \tau=\frac{\rho_{\max}+1}{2},\\
        &\mathcal{O} = \kappa \big[\widehat{\mathbf{V}}, \widehat{\mathbf{U}}\big] \in \mathbb{R}^{2\times H\times W},
	\end{aligned} 
\end{equation}
where $\kappa$ is a small scaling coefficient controlling the deformation magnitude. 

The final differentiable sampling grid is thus obtained by
\begin{equation}
	\begin{aligned}
		\label{eq21}
		&\mathcal{G}^{*} = \mathrm{Clip}\big(\widetilde{\mathcal{G}} + \mathcal{O}, -1, 1\big),
	\end{aligned}
\end{equation}

The differentiable operator $\mathrm{grid\_sample}$$(\cdot)$ is implemented on $\Gamma$ for obtaining resampled feature descriptors
\begin{equation}
	\tilde{\Gamma}=\mathrm{grid\_sample}(\Gamma,\,\mathcal{G}^{*}),
	\label{eq:grid_sample}
\end{equation}
which adapts the effective spatial support of the convolution according
to the predicted receptive field geometry.
Intuitively, this resampling step reshapes the region ``seen'' by each
kernel so that larger predicted scales \((N_u,N_v)\) correspond to a
broader or more elongated receptive field, without explicitly learning
pixel-wise offsets.

Meanwhile, a filter (i.e., $\Xi_{(N_u,N_v)}\in \mathbb{R}^{C_{\mathrm{in}} \times N_u \times N_v \times C_{\mathrm{out}}}$) with length $N_u$ and width $N_v$ is constructed where the filter parameters are randomly generated, $C_{\mathrm{in}}$ and $C_{\mathrm{out}}$ denote the channel numbers of the input and output feature maps of the current ARF convolution layer, respectively. The resampled feature descriptors $\tilde{\Gamma}$ 
\begin{equation}
	\bar{\Gamma}
	= \Xi_{(N_u,N_v)}\otimes\tilde{\Gamma},
	\label{eq:kernel_bank}
\end{equation}
where $\tilde{\Gamma}\in\mathbb{R}^{C_{\mathrm{in}}\times H\times W}$, $\bar{\Gamma}\in\mathbb{R}^{C_{\mathrm{out}}\times H\times W}$, and $\otimes$ denotes convolution. For notational simplicity, the batch dimension is omitted here.


In parallel, two lightweight convolutional branches (i.e., $\Delta_1$ and $\Delta_2$) generate a
modulation map and a bias term directly from the input feature map $\Gamma$ as follows:
\begin{equation}
	\begin{aligned}
		&M = \mathrm{Tanh}\!\big(\mathrm{Conv}_{3\times3}(\mathrm{Sigmoid}(\mathrm{Conv}_{3\times3}(\Gamma)))\big),\\
		&B= \mathrm{Conv}_{3\times3}(
		\mathrm{Sigmoid}(\mathrm{Conv}_{3\times3}(\Gamma))),
		\label{eq:bias_map}
	\end{aligned}
\end{equation}
where $\mathrm{Tanh}(\cdot)$ denotes the hyperbolic tangent function. The final ARF output is formulated as
\begin{equation}
	\widetilde{Y}
	= \bar{\Gamma} \odot M + B,
	\label{eq:arf_output}
\end{equation}
where $\odot$ denotes element-wise multiplication. The modulation term enhances discriminative regions, while the bias term
stabilizes the adaptive filtering process. 

To further highlight discriminative channel-wise information, an efficient channel attention mechanism is introduced after the intermediate response to adaptively enhance informative feature channels, thereby further improving the quality of feature representation. The final output is defined as
\begin{equation}
	Y=\mathrm{Sigmoid}\big(\mathrm{Conv}(\mathrm{GAP}(\widetilde{Y}))\big)\odot \widetilde{Y},
	\label{eq:eca_output}
\end{equation}

The designed ARF strategy enables the designed networks to capture feature descriptors with adaptive receptive fields from the input images (encompassing both spatial and frequency-domain information).

\subsubsection{ARF-based Spatial-frequency Feature Extraction}
In this subsection, an ARF-based spatial-frequency feature extraction strategy is constructed for extracting  spatial and spectral perspectives complementary representations from each input image $I(x,y)\in\mathbb{R}^{C\times H\times W}$, where $C$ is the channel number, $(H,W)$ are the spatial dimensions, and $(x,y)$ is a point location in the Cartesian coordinate
system.


After the input image $I(x,y)$ passes through the ARF-based spatial feature extraction strategy, spatial domain feature descriptors $\Theta_s(I)$ can be obtained. In parallel, a channel-wise 2D discrete cosine transform (DCT) is performed on the input image $I(x,y)$ for obtaining each channel's image frequency feature information as follows
\begin{equation}
	\mathcal{F}^{(c)}(u,v)
	=\alpha(u)\alpha(v)
	\sum_{x=0}^{H-1}\sum_{y=0}^{W-1}
	I^{(c)}(x,y)\,\phi_u(x)\phi_v(y),
	\label{eq:dct}
\end{equation}
where
$\phi_u(x)=\cos\!\big(\tfrac{\pi(2x+1)u}{2H}\big)$, $\phi_v(y)=\cos\!\big(\tfrac{\pi(2y+1)v}{2W}\big)$,
$\alpha(\cdot)$ is the standard DCT normalization, and $c$ denotes the channel.

After obtaining the channel-wise frequency representation $\mathcal{F}$ by DCT, a learnable spectral mask template is introduced to adaptively acquire salient frequency information prior to the inverse transformation. Specifically, we define a compact spectral mask template, $\mathcal{M}_0 \in \mathbb{R}^{1\times C\times 16\times 16}$. This parameterization acts as a coarse, channel-wise frequency weighting pattern with few learnable parameters. The template is bilinearly resized to match the target frequency resolution, followed by a Sigmoid activation to ensure differentiability and constrain its values to $[0,1]$:
\begin{equation}
	\mathcal{M}=\mathbf{1}-\mathrm{Sigmoid}\!\big(\mathrm{Upsample}(\mathcal{M}_0;\,H,W)\big),
	\label{eq:mask1}
\end{equation}
where $\mathbf{1}$ denotes an all-ones tensor with the same size as $\mathcal{M}$. Based on the learned spectral mask, the salient frequency descriptor is obtained as
\begin{equation}
	\tilde{\mathcal{F}}=\mathcal{F}\odot\mathcal{M},
	\label{eq:mask2}
\end{equation}
where $\odot$ denotes element-wise multiplication. Then an inverse DCT (i.e., $\mathcal{D}^{-1}(\cdot)$) is utilized for obtaining salient frequency descriptors' cossreponding spatial feature descriptors
\begin{equation}
	\Omega_f=\mathcal{D}^{-1}(\tilde{\mathcal{F}})
	\in\mathbb{R}^{C\times H\times W}.
	\label{eq:idct}
\end{equation}
After the feature descriptors $\Omega_f$  pass through the ARF-based frequency feature extraction strategy, the frequency-domain feature descriptors $\Theta_f(I)$ are obtained.
Thus, feature descriptors $\Theta_s(I)$ preserve localized spatial details, whereas $\Theta_f(I)$ emphasize informative spectral structures. Both are fed into the following adaptive feature fusion module.

\subsubsection{Adaptive Spatial-frequency Feature Fusion}
After passing through the ARF-based spatial and frequency backbones, we obtain two branch-wise feature maps $\Theta_s(I)$, $\Theta_f(I)$$\in$$\mathbb{R}^{C\times H\times W}$. These two streams are complementary: the spatial pathway focuses on local texture and shape cues, while the frequency pathway emphasizes global structure and spectral patterns. Therefore, fusing them is crucial for constructing discriminative representations in fine-grained recognition with few samples.
They are first concatenated along the channel dimension:
\begin{equation}
	F_{\mathrm{cat}} = \mathrm{Concat}(\Theta_s(I), \Theta_f(I))
	\in \mathbb{R}^{2C\times H\times W}.
	\label{eq:cat}
\end{equation}

A lightweight convolutional subnetwork $\psi(\cdot)$, consisting of three
convolutional layers with LeakyReLU activations, maps $F_{\mathrm{cat}}$
to a two-channel attention score map:
\begin{equation}
	A = \psi(F_{\mathrm{cat}})
	\in \mathbb{R}^{2\times H\times W},
	\label{eq:att_score}
\end{equation}
where $A = [A_s, A_f]$ and the two channels correspond to the spatial
and frequency branches, respectively.

A channel-wise softmax is applied to obtain normalized attention weights:
\begin{equation}
	W_s = \frac{\exp(A_s)}{\exp(A_s)+\exp(A_f)},~W_f = \frac{\exp(A_f)}{\exp(A_s)+\exp(A_f)}.
	\label{eq:softmax}
\end{equation}

The final fused representation is computed as a pixel-wise convex
combination of the two branches:
\begin{equation}
	F = W_s \odot \Theta_s(I) + W_f \odot \Theta_f(I)
	\in \mathbb{R}^{C\times H\times W}.
	\label{eq:fuse}
\end{equation}
This fusion mechanism allows the network to dynamically emphasize
spatially informative regions while leveraging complementary spectral
cues from the frequency branch, yielding robust fine-grained
representations.

\subsection{ResNet12-based ARF-SFE Module}
\begin{figure*}[htbp]
	\centering
	\includegraphics[width=0.8\linewidth]{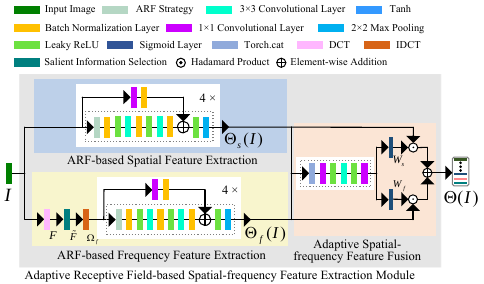}
	\caption{The architecture of ARF-SFR based on the ResNet-12 backbone.}
	\label{fig4}
\end{figure*}

Our designed architecture for ARF-SFE based on ResNet12~\cite{lee2019meta} is shown in Fig.~\ref{fig4}. It can be observed from Fig.~\ref{fig4} that our designed ARF strategy is employed for replacing the first convolutional layer in each module of the ResNet-12. It can also be observed that the proposed ResNet12-based ARF-SFE module has the capability to adaptively determine receptive field sizes for obtaining spatial and frequency features, and effectively fuse them for reconstruction and FSFGIC tasks.

\subsection{Similarity Measure Module}
After ARF-SFE encoding, feature descriptors of support sample $\boldsymbol{s}$ and query sample
$\boldsymbol{q}$ are represented as
$\Theta(\boldsymbol{s}), \Theta(\boldsymbol{q}) \in \mathbb{R}^{(H\times W)\times C}$, respectively,
which are reshaped into $m = H\times W$ tokens.

For an $N$-way $K$-shot episode, let $r\in\{1,\dots,N\}$ denote the class index.
The $K$ support samples of class $r$ are concatenated along the token dimension:
\begin{equation}
	\tilde{\boldsymbol{s}}_{r}
	= \operatorname{Concat}\big(
	\Theta(\tilde{\boldsymbol{s}}_{(r,1)}),\dots,
	\Theta(\tilde{\boldsymbol{s}}_{(r,K)})
	\big)
	\in \mathbb{R}^{(Km)\times C},
	\label{eq:s_concat}
\end{equation}
while the query tokens are
$\tilde{\boldsymbol{q}}\in \mathbb{R}^{m\times C}$. Both sets are projected into Q/K/V embeddings:
\begin{equation}
	\tilde{\boldsymbol{s}}_{r}^{Q,K,V}\!\in \mathbb{R}^{(Km)\times d_g},
	\qquad
	\tilde{\boldsymbol{q}}^{Q,K,V}\!\in \mathbb{R}^{m\times d_g},
	\label{eq:qkv_proj}
\end{equation}
where $d_g$ is the projection dimension. The correspondence between the query $\boldsymbol{q}$ and class $r$ is obtained via
scaled dot-product attention:
\begin{equation}
	\hat{\boldsymbol{q}}_{r}
	= \mathrm{Softmax}\!\left(
	\frac{
		\tilde{\boldsymbol{q}}^{Q}
		(\tilde{\boldsymbol{s}}_{r}^{K})^{\top}}
	{\sqrt{d_g}}
	\right)
	\tilde{\boldsymbol{s}}_{r}^{V},
	\label{eq:attn_s2q}
\end{equation}
and the support-side reconstruction $\hat{\boldsymbol{s}}_{r}$ is computed
symmetrically. The two alignment discrepancies are defined as
\begin{equation}
	\begin{aligned}
		&d_{\tilde{\boldsymbol{q}}\rightarrow\tilde{\boldsymbol{s}}_r}
		= \|\tilde{\boldsymbol{q}}^{V}-\hat{\boldsymbol{q}}_{r}\|_F^{2},\\
		&d_{\tilde{\boldsymbol{s}}_r\rightarrow\tilde{\boldsymbol{q}}}
		= \|\tilde{\boldsymbol{s}}_{r}^{V}-\hat{\boldsymbol{s}}_{r}\|_F^{2},
		\label{eq:bidist}
	\end{aligned}
\end{equation}
and combined into the class-wise dissimilarity
\begin{equation}
	d_r = \lambda_1d_{\tilde{\boldsymbol{q}}\rightarrow\tilde{\boldsymbol{s}}_r}
	+ \lambda_2d_{\tilde{\boldsymbol{s}}_r\rightarrow\tilde{\boldsymbol{q}}},
	\label{eq:dr}
\end{equation}
where $\lambda_1,\lambda_2$ are learnable weights. The prediction is obtained via a softmax over $\{-d_r\}$:
\begin{equation}
	P(y=r\mid\tilde{\boldsymbol{q}})
	= \frac{\exp(-d_r)}{\sum_{j=1}^{N}\exp(-d_j)},
	\label{eq:prob}
\end{equation}
and the entire model is trained with cross-entropy loss.

\section{Experiments}

\subsection{Datasets}

The proposed ARF-SFR-Net is evaluated on five widely-used FSFGIC benchmarks:~meta-iNat~\cite{wertheimer2019few}, tiered meta-iNat~\cite{wertheimer2019few}, CUB-200-2011~\cite{wah2011caltech}, Stanford Dogs~\cite{khosla2011novel}, and Stanford Cars~\cite{krause20133d}. The meta-iNat dataset consists of 1,135 wildlife categories. The tiered meta-iNat dataset is designed with a larger domain shift between base and novel classes to assess model generalization under a more rigorous scenario. The CUB-200-2011 dataset encompasses 200 bird species with a total of 11,788 images. The Stanford Dogs dataset includes 20,580 images across 120 dog breeds. The Stanford Cars dataset contains 16,185 images of cars from 196 categories. For a fair comparison with established works, we strictly adhere to the standard dataset splits as detailed in Table~\ref{tab:1}.
\begin{table}[htbp]
	\caption{The class split of the five datasets. $N_{\text{train}}$, $N_{\text{val}}$, and $N_{\text{test}}$ denote the numbers of classes in the training, validation, and test sets respectively.}
	\label{tab:1}
	\renewcommand{\arraystretch}{0.9}
	\setlength{\tabcolsep}{4pt}
	\centering
	\scriptsize
	\begin{tabular}{lccc}
		\toprule
		Dataset & $N_{\text{train}}$ & $N_{\text{val}}$ & $N_{\text{test}}$ \\
		\midrule
		meta-iNat & 908 & - & 227 \\
		tiered meta-iNat & 781 & - & 354\\
		CUB-200-2011 & 100 & 50 & 50 \\
		Stanford Dogs & 70 & 20 & 30 \\
		Stanford Cars & 130 & 17 & 49 \\
		\bottomrule
	\end{tabular}
\end{table}

\subsection{Experimental Setup}

We conducted experiments on the aforementioned five datasets under the 5-way 1-shot and 5-way 5-shot FSFGIC settings. The proposed ARF-SFR-Net is trained from scratch in an end-to-end manner, with no fine-tuning applied during testing. During training, the parameters in different convolutional blocks of ARF-SFR-Net are not shared. Each input image is resized to a fixed resolution of 92$\times$92 and randomly cropped to 84$\times$84.

In our experiments, models based on Conv-4~\cite{snell20177} and ResNet-12~\cite{lee2019meta} are trained for 1,200 epochs using SGD with Nesterov momentum (0.9) and a weight decay of \(5 \times 10^{-4}\). The learning rate is initialized at 0.1 and reduced by a factor of 10 every 400 epochs. Conv-4 models are trained using 30-way 5-shot episodes, while ResNet-12 models used 15-way 5-shot episodes to reduce memory usage. All models are evaluated on the standard 5-way 1-shot and 5-shot tasks. Validation is performed every 20 epochs, and the checkpoint with the best validation performance is retained for final testing. Reported results correspond to the mean accuracy over 10,000 randomly sampled test episodes under 5-way 1-shot and 5-shot settings. The designed ARF strategy is employed in all four convolutional layers of Conv-4~\cite{snell20177} and the last two blocks of ResNet-12~\cite{lee2019meta} (see Ablation study).

For snapshot training, we use a cosine annealing learning rate schedule, saving a model snapshot at each minimum (every 240 epochs) to create a 5-model ensemble for validation-based selection. During inference, the final prediction is obtained by averaging the output probabilities of all snapshots, and the mean accuracy and variance on the test set are reported.
\begin{table}[htbp]
	\caption{Comparison results of different methods on the meta-iNat and tiered meta-iNat datasets in the 5-way setting under the Conv-4 backbone. The best performance is indicated in bold.}
	\renewcommand{\arraystretch}{0.8}
	\setlength{\tabcolsep}{6pt}
	\centering
	\scriptsize
	\sisetup{table-format=2.2} 
	\begin{tabular}{
			l
			S[table-format=1]
			S[table-format=1]
			S[table-format=1]
			S[table-format=1]
		}
		\toprule[1.pt]
		\multirow{2}{*}{Method} & \multicolumn{2}{c}{meta-iNat} & \multicolumn{2}{c}{tiered meta-iNat} \\
		\cmidrule(lr){2-3} \cmidrule(lr){4-5}
		& {1-shot} & {5-shot} & {1-shot} & {5-shot} \\
		\midrule
		ProtoNet~\cite{snell20177} & 53.78 & 73.80 & 35.47 & 54.85 \\
		DSN~\cite{ruan2021few} & 58.08 & 77.38 & 36.82 & 60.11 \\
		FRN~\cite{Wertheimer2021FRN} & 61.98 & 80.04 & 43.95 & 63.45 \\
		FRN+TDM~\cite{lee2022task} & 63.97 & 81.60 & 44.05 & 62.91 \\
		DeepEMD~\cite{zhang2022deepemd} & 54.48 & 68.36 & 36.05 & 48.55 \\
		C2-Net~\cite{ma2024cross} & 71.47 & 85.47 & 49.04 & 67.25 \\
		TST\_MFL~\cite{sun2025tst_mfl} & 69.71 & 81.04 & 45.45 & 62.12 \\
		TDM+IAM~\cite{lee2024task} & 65.95 & 83.30 & 46.45 & 66.55 \\
		AIS~\cite{zhao2024angular} & 68.40 & 82.46 & 48.97 & 67.15 \\
		BDFRNet~\cite{wu2024bi1} & 66.07 & 83.30 & 46.64 & 66.46 \\
		BTG-Net~\cite{ma2024bi} & 71.36 & 84.59 & 50.62 & 69.11 \\
		SUITED~\cite{ma2025few} & 74.72 & 87.44 &  51.70 & 70.43 \\
		\midrule
		Ours & \bfseries \textbf{75.21} & \bfseries \textbf{88.95} & ~\bfseries \textbf{52.78} & ~\bfseries \textbf{73.67} \\
		Ours-Snapshot & \bfseries \textbf{76.11} & \bfseries \textbf{89.90} & ~\bfseries \textbf{53.95} & ~\bfseries \textbf{75.08} \\
		
		\bottomrule[1.pt]
	\end{tabular}
	\label{t3}
\end{table}

\subsection{Performance Comparison}

This section presents a comparative analysis between the proposed ARF-SFR-Net and twenty-nine state-of-the-art methods. As summarized in Table~\ref{t3} and Table~\ref{t2}, our method achieves the allover best performance on five evaluated datasets. For instance, on the CUB-200-2011 dataset using a Conv-4 backbone, ARF-SFR-Net achieves accuracies of 81.84$\%$ (5-way 1-shot) and 93.24$\%$ (5-way 5-shot), outperforming the current best alternative, C2-Net (78.63$\%$ and 89.48$\%$, respectively)~\cite{ma2024cross}. These results validate the efficacy of our designed spatial–frequency feature extraction with an adaptive receptive field architecture.

\begin{figure}[h]
	\centering
	\includegraphics[width=0.8\linewidth]{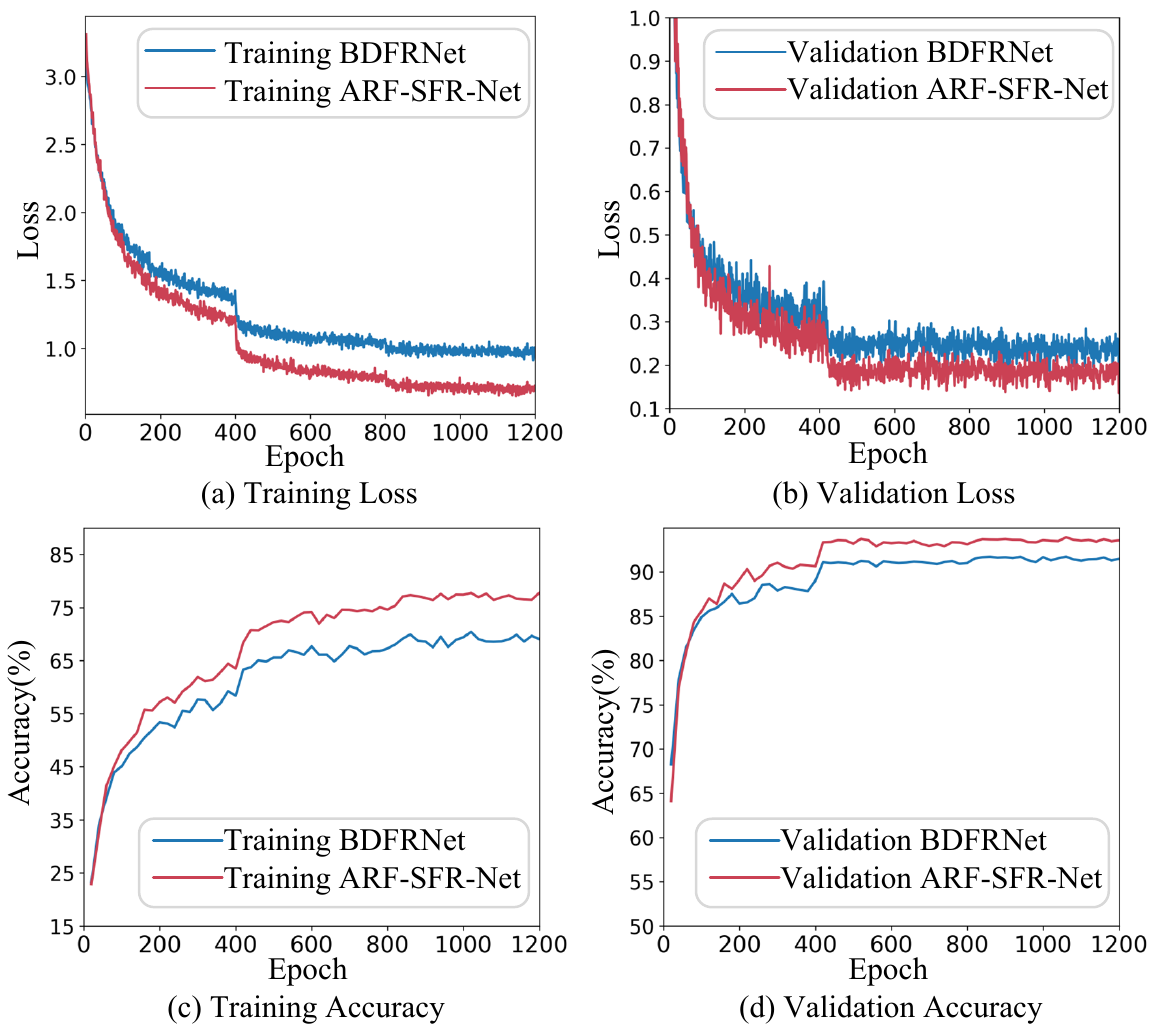}
	\caption{Examples of the loss and accuracy curves of BDFRNet and the proposed method for the 5-way 1-shot FSFGIC task on the CUB-200-2011 dataset.}
	\label{fig:3}
\end{figure}

Taking the 5-way 1-shot task on the CUB-200-2011 dataset as an example, Fig.~\ref{fig:3} compares the training and validation performance of BDFRNet and the proposed ARF-SFR-Net. The loss curves (Fig.~\ref{fig:3}(a) and (b)) and accuracy curves (Fig.~\ref{fig:3}(c) and (d)) demonstrate that ARF-SFR-Net achieves consistently lower loss and higher accuracy than BDFRNet throughout both training and validation phases.
\begin{table*}[!htbp]
	\caption{Comparison results of different methods for 5-way tasks on the CUB-200-2011, Stanford Dogs, and Stanford Cars datasets with two different backbones. The best performance is indicated in bold.}
	\renewcommand{\arraystretch}{1}
	\centering
	\resizebox{\textwidth}{!}{
		\begin{tabular}{@{}c|c|cc|cc|cc@{}}
			\toprule
			\multirow{2}{*}{Backbone} & \multirow{2}{*}{Method} & \multicolumn{2}{c|}{CUB-200-2011}          & \multicolumn{2}{c|}{Stanford Dogs}    & \multicolumn{2}{c}{Stanford Cars}    \\ 
			\cline{3-8}
			&  & 1-shot & 5-shot & 1-shot & 5-shot  & 1-shot & 5-shot \\ 
			\cline{1-8}
			\multirow{18}{*}{Conv-4} 
			& DSN~\cite{ruan2021few}       & 72.56$\pm$0.92      & 84.62$\pm$0.60     & 44.52$\pm$0.82   & 59.42$\pm$0.71   & 53.45$\pm$0.86   & 65.19$\pm$0.75   \\
			& LRPABN~\cite{huang2021low}      & 63.63$\pm$0.77      & 76.06$\pm$0.58     & 45.72$\pm$0.75   & 60.94$\pm$0.66   & 60.28$\pm$0.76   & 73.29$\pm$0.58   \\
			& FRN~\cite{Wertheimer2021FRN}       & 74.90$\pm$0.21      & 89.39$\pm$0.12     & 60.41$\pm$0.21   & 79.26$\pm$0.15   & 67.48$\pm$0.22   & 87.97$\pm$0.11   \\
			& FRN+TDM~\cite{lee2022task}   & 74.39$\pm$0.21      & 88.89$\pm$0.11     & 62.77$\pm$0.22   & 79.71$\pm$0.14   & 72.26$\pm$0.21   & 89.55$\pm$0.10   \\
			& PaCL~\cite{wang2022pacl}      & 74.07$\pm$0.70      & 88.75$\pm$0.38     & 59.76$\pm$0.70   & 77.50$\pm$0.48   & 72.21$\pm$0.68   & 88.02$\pm$0.36   \\
			& DAN~\cite{Shulin2022}     & 72.89$\pm$0.50      & 86.60$\pm$0.31         & 59.81$\pm$0.50   & 77.19$\pm$0.35 & 70.21$\pm$0.50   & 85.55$\pm$0.31  \\
			& DeepEMD~\cite{zhang2022deepemd}    & 64.08$\pm$0.50      & 80.55$\pm$0.71     & 46.73$\pm$0.49   & 65.74$\pm$0.63   & 61.63$\pm$0.27   & 72.95$\pm$0.38   \\
			
			
			& ATR-Net~\cite{yu2024adaptive} & 75.24$\pm$0.46      & 87.25$\pm$0.29     &  61.17$\pm$0.50   &  76.57$\pm$0.35   &  66.07$\pm$0.47   &  82.39$\pm$0.31  \\
			&AIS~\cite{zhao2024angular}& 79.56      & 89.92      & 63.13$\pm$0.51     & 78.34$\pm$0.35       &-     & -    \\
			& FRN+CSCAM~\cite{yang2024channel} & 77.68$\pm$0.20      & 89.88$\pm$0.12     & -   & -   & 71.44$\pm$0.21   & 86.44$\pm$0.13  \\
			&TDM+CSCAM~\cite{yang2024channel}& 77.81$\pm$0.21      &  89.60$\pm$0.12     &-     & -     & 73.27$\pm$0.21    & 87.81$\pm$0.12     \\
			&SUITED~\cite{ma2025few}& 79.73      &  90.05     & 68.67$\pm$0.51    &82.24$\pm$0.32      &82.21$\pm$0.44     & 92.39$\pm$0.24    \\
			& BDFRNet$\dagger$~\cite{wu2024bi1}     & 76.39$\pm$0.20      & 90.61$\pm$0.11     & 64.66$\pm$0.22   & 81.27$\pm$0.14   & 75.33$\pm$0.20   & 90.91$\pm$0.10   \\
			& C2-Net$\dagger$~\cite{ma2024cross} & 78.63$\pm$0.46      & 89.48$\pm$0.26     & 69.81$\pm$0.50   & 84.39$\pm$0.29   & 79.52$\pm$0.45   & 91.15$\pm$0.24   \\
			& BinoHeM~\cite{qi2025binohem}      & 72.18$\pm$0.72      & 85.85$\pm$0.45     & -   & -   & 51.53$\pm$0.66   & 70.62$\pm$0.52   \\
			\cline{2-8}
			& Ours           & \bfseries 80.95$\pm$0.19 & \bfseries 92.84$\pm$0.10 & \bfseries 72.26$\pm$0.21 & \bfseries 86.68$\pm$0.12 & \bfseries 85.24$\pm$0.16 & \bfseries 95.59$\pm$0.06 \\
			& Ours-Snapshot            & \bfseries 83.20$\pm$0.18 & \bfseries 93.81$\pm$0.09 & \bfseries 75.45$\pm$0.20 & \bfseries 88.87$\pm$0.11 & \bfseries 84.90$\pm$0.17 & \bfseries 95.51$\pm$0.06   \\ \midrule
			\multirow{20}{*}{ResNet-12}
			& FRN~\cite{Wertheimer2021FRN}       & 82.86$\pm$0.19      & 92.41$\pm$0.11     & 76.76$\pm$0.21   & 88.74$\pm$0.12   & 86.90$\pm$0.17   & 95.69$\pm$0.07   \\
			& FRN+TDM~\cite{lee2022task}   & 83.26$\pm$0.20      & 92.80$\pm$0.11     & 75.98$\pm$0.22   & 88.70$\pm$0.13   & 86.91$\pm$0.17   & 96.11$\pm$0.07   \\
			& HelixFormer~\cite{zhang2022crossimage}   & 81.66$\pm$0.30     & 91.83$\pm$0.17     & 65.92$\pm$0.49   & 80.65$\pm$0.36   & 79.40$\pm$0.43   & 92.26$\pm$0.15   \\
			& DeepBDC~\cite{xie2022joint}   &  81.98$\pm$0.44     & 92.24$\pm$0.24     & 73.57$\pm$0.46   &  86.61$\pm$0.27   & 82.28$\pm$0.42   & 93.51$\pm$0.20   \\
			& BSFANet~\cite{zha2023boosting}      & 82.27$\pm$0.46      & 90.76$\pm$0.26     & 69.58$\pm$0.50   & 82.59$\pm$0.33   & 88.93$\pm$0.38   & 95.20$\pm$0.20   \\
			& SRNet~\cite{li2024self}      & 83.82$\pm$0.18      & 93.45$\pm$0.10     & 76.54$\pm$0.21   & 88.52$\pm$0.12   & 88.02$\pm$0.16   & 96.23$\pm$0.07   \\
			
			& ATR-Net~\cite{yu2024adaptive} &  81.37$\pm$0.42      & 91.26$\pm$0.24     &   75.68$\pm$0.46   &   87.77$\pm$0.26   &  85.60$\pm$0.38   &  95.24$\pm$0.16  \\
			& FRN+TDM+BiFI-TDM~\cite{li2024rise} & 83.80$\pm$0.19      &  94.26$\pm$0.09     &   76.72$\pm$0.20   &   88.96$\pm$0.12   &   89.15$\pm$0.16   &   97.31$\pm$0.05  \\
			& TST$\_$MFL~\cite{sun2025tst_mfl}  & 80.79$\pm$0.84      &  91.15$\pm$0.40     &   67.84$\pm$0.82   &   81.72$\pm$0.61   &   82.82$\pm$0.84   &   93.61$\pm$0.35  \\
			&AIS~\cite{zhao2024angular}& 85.60$\pm$0.42      &  93.36$\pm$0.21     & 76.32$\pm$0.47     & 88.25$\pm$0.27      & -    & -    \\
			& FRN+CSCAM~\cite{yang2024channel} & 84.00$\pm$0.18      & 93.52$\pm$0.10     & -   & -   & 86.24$\pm$0.18   & 95.55$\pm$0.08  \\
			&TDM+CSCAM~\cite{yang2024channel}& 83.34$\pm$0.19      &  92.98$\pm$0.10     &-     & -     & 86.86$\pm$0.17    & 95.63$\pm$0.08     \\
			&SUITED~\cite{ma2025few}& 86.02      &  94.13     & 76.55$\pm$0.47    &88.86$\pm$0.27      & 89.90 $\pm$0.36    &96.53$\pm$0.16     \\
			& BDFRNet$\dagger$~\cite{wu2024bi1}     & 82.03$\pm$0.19      & 92.78$\pm$0.10     & 77.40$\pm$0.21   & 88.41$\pm$0.12   & 90.28$\pm$0.14   & 97.26$\pm$0.05   \\
			& C2-Net$\dagger$~\cite{ma2024cross}      & 83.65$\pm$0.20      & 92.57$\pm$0.10     & 77.72$\pm$0.46   & 89.59$\pm$0.24   & 86.48$\pm$0.40   & 94.07$\pm$0.22   \\
			& FCAM~\cite{tang2026harnessing}      & 82.89$\pm$0.27      & 93.06$\pm$0.39     & -   & -   & -   & -   \\
			& MFSI-Net~\cite{yu2025multiscale}      & 80.70$\pm$0.42      & 91.30$\pm$0.24     & 76.35$\pm$0.46   & 88.17$\pm$0.26   & 85.60$\pm$0.37   & 95.26$\pm$0.16   \\
			
			\cline{2-8}
			& Ours            & \bfseries 86.08$\pm$0.17 & \bfseries 94.83$\pm$0.08 & \bfseries 78.65$\pm$0.20 & \bfseries 90.20$\pm$0.10 & \bfseries 91.24$\pm$0.13 & \bfseries 97.62$\pm$0.05 \\
			& Ours-Snapshot           & \bfseries 86.96$\pm$0.17 & \bfseries 95.23$\pm$0.08 & \bfseries 80.93$\pm$0.19 & \bfseries 90.68$\pm$0.11 & \bfseries 92.62$\pm$0.12 & \bfseries 98.15$\pm$0.04 \\ 
			\bottomrule[1.15pt]
	\end{tabular}}
	\label{t2} 
\end{table*}

\subsection{Ablation Study}
To further study the sensitivity of our approach, ablation experiments are conducted on the CUB-200-2011, Stanford Dogs, and Stanford Cars datasets as follows.

\textbf{ARF Strategy Applied on Conv-4.} We evaluate the impact of the ARF strategy applied to different convolutional blocks of Conv-4 (i.e., the first and second blocks (named as $\text{ARF}_{{\text{Conv4}}_{12}}$), the second and third blocks (named as $\text{ARF}_{{\text{Conv4}}_{23}}$), the third and fourth blocks (named as $\text{ARF}_{{\text{Conv4}}_{34}}$), and four blocks (named as $\text{ARF}_{{\text{Conv4}}_{1-4}}$)) on performing 5-way 1-shot and 5-way 5-shot FSFGIC tasks using the CUB-200-2011, Stanford Dogs, and Stanford Cars datasets. Results are presented in Table~\ref{t7}. It can be observed that the ARF strategy applied to the four blocks (i.e., $\text{ARF}_{{\text{Conv4}}_{1-4}}$) achieves the best overall performance on the three datasets, and is therefore adopted as the optimal configuration.

\begin{table}[h]
	\centering
	\scriptsize
	\caption{ARF strategy applied on Conv-4 for 5-way tasks on the CUB-200-2011, Stanford Dogs, and Stanford Cars datasets.}
	\renewcommand{\arraystretch}{0.95}
	\setlength{\tabcolsep}{2pt}
	\begin{tabular}{l c c c c c c}
		\toprule
		\multirow{2}{*}{Model} & \multicolumn{2}{c}{CUB-200-2011} & \multicolumn{2}{c}{Stanford Dogs} & \multicolumn{2}{c}{Stanford Cars} \\
		\cmidrule(lr){2-3} \cmidrule(lr){4-5} \cmidrule(lr){6-7}
		& 1-shot & 5-shot & 1-shot & 5-shot & 1-shot & 5-shot \\
		\hline
		$\text{ARF}_{{\text{Conv4}}_{12}}$ & 80.28$\pm$0.20 & 92.36$\pm$0.11 & 69.78$\pm$0.22 & 85.18$\pm$0.13 & 82.11$\pm$0.18 & 94.53$\pm$0.08 \\
		$\text{ARF}_{{\text{Conv4}}_{23}}$ & 79.96$\pm$0.21 & 92.08$\pm$0.12 & 69.41$\pm$0.23 & 84.86$\pm$0.14 & 81.76$\pm$0.19 & 94.22$\pm$0.09 \\
		$\text{ARF}_{{\text{Conv4}}_{34}}$ & 79.83$\pm$0.22 & 91.94$\pm$0.13 & 69.27$\pm$0.24 & 84.73$\pm$0.15 & 81.63$\pm$0.20 & 94.10$\pm$0.10 \\
		$\text{ARF}_{{\text{Conv4}}_{1-4}}$ & \bfseries 80.95$\pm$0.19 & \bfseries 92.84$\pm$0.10 & \bfseries 72.26$\pm$0.21 & \bfseries 86.68$\pm$0.12 & \bfseries 85.24$\pm$0.16 & \bfseries 95.59$\pm$0.06 \\
		\bottomrule
	\end{tabular}
	\label{t7}
\end{table}


\textbf{ARF Strategy Applied on ResNet-12.} We first evaluate the impact of the ARF strategy applied to different convolutional layers in different blocks of ResNet-12 (i.e., the first layer of each block (named as $\text{ARF}_{{\text{ResNet}}_{1}}$, the second layer of each block (named as $\text{ARF}_{{\text{ResNet}}_{2}}$, and the third layer of each block (named as $\text{ARF}_{{\text{ResNet}}_{3}}$) on performing 5-way 1-shot and 5-way 5-shot FSFGIC tasks using the CUB-200-2011, Stanford Dogs, and Stanford Cars datasets. Results show that applying ARF to the second layer of each block (i.e., $\text{ARF}_{{\text{ResNet}}_{2}}$) performs best. We then examined ARF on the second convolutional layer of different block pairs in ResNet-12 (i.e., the first and second blocks (named as $\text{ARF}_{{\text{ResNet}}_{12}}$), the second and third blocks (named as $\text{ARF}_{{\text{ResNet}}_{23}}$), and the third and fourth blocks (named as $\text{ARF}_{{\text{ResNet}}_{34}}$)) on performing 5-way 1-shot and 5-way 5-shot FSFGIC tasks using the CUB-200-2011, Stanford Dogs, and Stanford Cars datasets. It can be observed from Table~\ref{t8} that applying ARF to the second layer of the third and fourth blocks (i.e., $\text{ARF}_{{\text{ResNet}}_{34}}$) achieves the highest overall performance across the three datasets and is selected as the optimal configuration.

\begin{table}[h]
	\centering
	\scriptsize
	\caption{ARF strategy applied on ResNet-12 for 5-way tasks on the CUB-200-2011, Stanford Dogs, and Stanford Cars datasets.}
	\renewcommand{\arraystretch}{0.95}
	\setlength{\tabcolsep}{0.8pt}
	\begin{tabular}{l c c c c c c}
		\toprule
		\multirow{2}{*}{Model} & \multicolumn{2}{c}{CUB-200-2011} & \multicolumn{2}{c}{Stanford Dogs} & \multicolumn{2}{c}{Stanford Cars} \\
		\cmidrule(lr){2-3} \cmidrule(lr){4-5} \cmidrule(lr){6-7}
		& 1-shot & 5-shot & 1-shot & 5-shot & 1-shot & 5-shot \\
		\hline
		$ARF_{{\text{ResNet}}_{1}}$   & 83.92$\pm$0.20 & 93.41$\pm$0.09 & 76.21$\pm$0.22 & 88.64$\pm$0.12 & 88.87$\pm$0.15 & 96.64$\pm$0.07 \\
		$\text{ARF}_{{\text{ResNet}}_{2}}$  & \bfseries 86.11$\pm$0.15 & \bfseries 94.87$\pm$0.17 & \bfseries 78.68$\pm$0.22 &  90.12$\pm$0.11 &  91.22$\pm$0.14 &  97.70$\pm$0.06  \\
		$\text{ARF}_{{\text{ResNet}}_{3}}$   & 84.57$\pm$0.19 & 94.05$\pm$0.08 & 77.14$\pm$0.21 & 89.31$\pm$0.10 & 89.86$\pm$0.14 & 97.02$\pm$0.06 \\
		\hline
		$\text{ARF}_{{\text{ResNet}}_{12}}$   & 86.10$\pm$0.18 & 94.42$\pm$0.08 & 78.16$\pm$0.20 & 90.16$\pm$0.12 & 91.02$\pm$0.13 & 97.28$\pm$0.05 \\
		$\text{ARF}_{{\text{ResNet}}_{23}}$   & 85.88$\pm$0.18 & 94.57$\pm$0.08 & 78.35$\pm$0.20 & 90.08$\pm$0.17 & 91.15$\pm$0.13 & 97.11$\pm$0.05 \\
		$\text{ARF}_{{\text{ResNet}}_{34}}$   &   86.08$\pm$0.17 &  94.83$\pm$0.08 & 78.65$\pm$0.20 & \bfseries 90.20$\pm$0.10 & \bfseries 91.24$\pm$0.13 & \bfseries 97.62$\pm$0.05 \\
		\bottomrule
	\end{tabular}
	\label{t8}
\end{table}


\textbf{Impact of ARF-based feature descriptor extraction in the spatial and frequency domains on the accuracy of FSFGIC.} We evaluate the impact of ARF-based feature descriptor extraction in the spatial and frequency domains on performance (i.e., extracting feature descriptors only in the spatial domain (named as ARF-SFR-Net$\_{S}$), extracting feature descriptors only in the frequency domain (named as ARF-SFR-Net$\_{F}$),  extracting feature descriptors in the spatial and frequency domains (named as ARF-SFR-Net)) on performing 5-way 1-shot and 5-way 5-shot FSFGIC tasks using the CUB-200-2011, Stanford Dogs, and Stanford Cars datasets on Conv-4 and ResNet-12 backbones. It can be found from Table~\ref{t4} that ARF-based feature descriptor extraction in the spatial and frequency domains together (i.e., ARF-SFR-Net) achieves the best performance on Conv-4 and ResNet-12 backbones, and is therefore adopted as the optimal configuration.

\begin{table}[!htbp]
	\caption{Impact of ARF-based feature descriptor extraction in the spatial and frequency domains for 5-way tasks on the CUB-200-2011, Stanford Dogs, and Stanford Cars datasets.}
	\label{tab:domain-performance}
	\centering
	\renewcommand{\arraystretch}{1.0}
	\setlength{\tabcolsep}{1.5pt}
	
	\resizebox{\linewidth}{!}{%
		\begin{tabular}{c c cc cc cc}
			\toprule
			\multirow{2}{*}{Model} & \multirow{2}{*}{Backbone} &
			\multicolumn{2}{c}{CUB-200-2011} &
			\multicolumn{2}{c}{Stanford Dogs} &
			\multicolumn{2}{c}{Stanford Cars} \\
			\cmidrule(lr){3-4} \cmidrule(lr){5-6} \cmidrule(lr){7-8}
			& & 1-shot & 5-shot & 1-shot & 5-shot & 1-shot & 5-shot \\
			\midrule
			ARF-SFR-Net$_S$ & Conv-4 &
			78.33 $\pm$ 0.20 & 91.06 $\pm$ 0.11 &
			69.01 $\pm$ 0.21 & 85.41 $\pm$ 0.13 &
			81.21 $\pm$ 0.18 & 94.50 $\pm$ 0.07 \\
			ARF-SFR-Net$_F$ & Conv-4 &
			78.40 $\pm$ 0.23 & 92.20 $\pm$ 0.16 &
			68.22 $\pm$ 0.21 & 82.58 $\pm$ 0.17 &
			78.97 $\pm$ 0.22 & 94.71 $\pm$ 0.15 \\
			ARF-SFR-Net & Conv-4 &
			\textbf{80.95 $\pm$ 0.19} & \textbf{92.84 $\pm$ 0.10} &
			\textbf{72.26 $\pm$ 0.21} & \textbf{86.68 $\pm$ 0.12} &
			\textbf{85.24 $\pm$ 0.16} & \textbf{95.59 $\pm$ 0.06} \\
			\midrule
			ARF-SFR-Net$_S$ & ResNet-12 &
			83.31 $\pm$ 0.19 & 93.47 $\pm$ 0.09 &
			77.64 $\pm$ 0.21 & 88.87 $\pm$ 0.12 &
			90.90 $\pm$ 0.14 & 97.57 $\pm$ 0.05 \\
			ARF-SFR-Net$_F$ & ResNet-12 &
			83.55 $\pm$ 0.18 & 93.39 $\pm$ 0.09 &
			76.98 $\pm$ 0.21 & 89.03 $\pm$ 0.12 &
			87.91 $\pm$ 0.15 & 96.88 $\pm$ 0.03 \\
			ARF-SFR-Net & ResNet-12 &
			\textbf{86.08 $\pm$ 0.17} & \textbf{94.83 $\pm$ 0.08} &
			\textbf{78.65 $\pm$ 0.20} & \textbf{90.20 $\pm$ 0.10} &
			\textbf{91.24 $\pm$ 0.13} & \textbf{97.62 $\pm$ 0.05} \\
			\bottomrule
		\end{tabular}
	}
	\label{t4}
\end{table}

\textbf{Impact of ARF strategy for FSFGIC.} We evaluate the impact of ARF strategy (i.e., the designed network without AFR strategy (named as SFR-Net) and the designed network with ARF (named as ARF-SFR-Net)) on performing 5-way 1-shot and 5-way 5-shot FSFGIC tasks using the CUB-200-2011, Stanford Dogs, and Stanford Cars datasets on Conv-4 and ResNet-12 backbones. It can be found from Table~\ref{t5} that the designed network with ARF (i.e., ARF-SFR-Net) achieves the best performance on Conv-4 and ResNet-12 backbones.


\textbf{Impact of the maximum selectable kernel size $\rho_{\max}$ on FSFGIC.} We evaluate the impact of different maximum selectable kernel sizes (i.e., $\rho_{\max}\in\{3,5,7,9,11\}$) on performing 5-way 1-shot and 5-way 5-shot FSFGIC tasks using the CUB-200-2011, Stanford Dogs, and Stanford Cars datasets on Conv-4 and ResNet-12 backbones. Results are presented in Table~\ref{t}. It can be found from Table~\ref{t6} that the design ARF-SFR-Net with ${\rho={9}}$ achieves the best overall performance on the Conv-4 and ResNet-12 backbones, and is therefore adopted as the optimal configuration.

\begin{table*}[!htbp]
	\centering
	\caption{The influence of ARF strategy on the proposed ARF-SFR-Net with Conv-4 and ResNet12 using the CUB-200-2011, Stanford Dogs, and Stanford Cars datasets for 5-way tasks.}
	\renewcommand{\arraystretch}{0.9}
	
	\resizebox{\linewidth}{!}{%
		{\begin{tabular}{c c cc cc cc}
				
				\toprule
				\multirow{2}{*}{Model} &  \multirow{2}{*}{Backbone}  & \multicolumn{2}{c}{CUB-200-2011} & \multicolumn{2}{c}{Stanford Dogs} &  \multicolumn{2}{c}{Stanford Cars}\\
				\cmidrule(lr){3-4} \cmidrule(lr){5-6} \cmidrule(lr){7-8}
				&  &  1-shot & 5-shot & 1-shot & 5-shot  & 1-shot & 5-shot \\
				\cline{1-8}
				
				
				SFR-Net & Conv-4  &78.33$\pm$0.20 &92.06$\pm$0.11   &68.01$\pm$0.23 &83.01$\pm$0.15 &80.21$\pm$0.12 &93.15$\pm$0.10 \\
				ARF-SFR-Net & Conv-4 &
				\textbf{80.95 $\pm$ 0.19} & \textbf{92.84 $\pm$ 0.10} &
				\textbf{72.26 $\pm$ 0.21} & \textbf{86.68 $\pm$ 0.12} &
				\textbf{85.24 $\pm$ 0.16} & \textbf{95.59 $\pm$ 0.06} \\
				\midrule
				SFR-Net & ResNet-12 &
				83.31 $\pm$ 0.19 & 93.07 $\pm$ 0.09 &
				77.14 $\pm$ 0.21 & 88.77 $\pm$ 0.12 &
				90.30 $\pm$ 0.14 & 96.57 $\pm$ 0.05 \\
				ARF-SFR-Net & ResNet-12 &
				\textbf{86.08 $\pm$ 0.17} & \textbf{94.83 $\pm$ 0.08} &
				\textbf{78.65 $\pm$ 0.20} & \textbf{90.20 $\pm$ 0.10} &
				\textbf{91.24 $\pm$ 0.13} & \textbf{97.62 $\pm$ 0.05} \\
				
				\bottomrule
		\end{tabular}}
	}
	\label{t5}
\end{table*}

\begin{table}[!htbp]
	\caption{Impact of different $\rho_{\max}$ settings on the proposed ARF-SFR-Net for 5-way tasks on the CUB-200-2011, Stanford Dogs, and Stanford Cars datasets.}
	\centering
	\renewcommand{\arraystretch}{1.0}
	\setlength{\tabcolsep}{1.5pt}
	
	\resizebox{\linewidth}{!}{%
		\begin{tabular}{c c cc cc cc}
			\toprule
			\multirow{2}{*}{Model} & \multirow{2}{*}{Backbone} &
			\multicolumn{2}{c}{CUB-200-2011} &
			\multicolumn{2}{c}{Stanford Dogs} &
			\multicolumn{2}{c}{Stanford Cars} \\
			\cmidrule(lr){3-4} \cmidrule(lr){5-6} \cmidrule(lr){7-8}
			& & 1-shot & 5-shot & 1-shot & 5-shot & 1-shot & 5-shot \\
			\midrule
			ARF-SFR-Net$_{\rho_{\max}=3}$ & Conv-4 &
			78.86 $\pm$ 0.22 & 91.47 $\pm$ 0.12 &
			69.54 $\pm$ 0.23 & 84.93 $\pm$ 0.14 &
			82.18 $\pm$ 0.19 & 94.18 $\pm$ 0.08 \\
			ARF-SFR-Net$_{\rho_{\max}=5}$ & Conv-4 &
			79.63 $\pm$ 0.21 & 92.11 $\pm$ 0.11 &
			70.82 $\pm$ 0.22 & 85.76 $\pm$ 0.13 &
			83.47 $\pm$ 0.18 & 94.86 $\pm$ 0.07 \\
			ARF-SFR-Net$_{\rho_{\max}=7}$ & Conv-4 &
			80.37 $\pm$ 0.20 & 92.56 $\pm$ 0.10 &
			71.64 $\pm$ 0.21 & 86.21 $\pm$ 0.12 &
			84.53 $\pm$ 0.17 & 95.24 $\pm$ 0.06 \\
			ARF-SFR-Net$_{\rho_{\max}=9}$ & Conv-4 &
			\textbf{80.95 $\pm$ 0.19} & 92.84 $\pm$ 0.10 &
			\textbf{72.26 $\pm$ 0.21} & 86.68 $\pm$ 0.12 &
			\textbf{85.24 $\pm$ 0.16} & \textbf{95.59 $\pm$ 0.06} \\
	ARF-SFR-Net$_{\rho_{\max}=11}$ & Conv-4 &
		   80.90 $\pm$ 0.22 & \textbf{92.89 $\pm$ 0.12} &
			72.11 $\pm$ 0.24 & \textbf{86.81 $\pm$ 0.15} &
			85.20 $\pm$ 0.21 & 95.51 $\pm$ 0.11 \\
			\midrule
			ARF-SFR-Net$_{\rho_{\max}=3}$ & ResNet-12 &
			84.21 $\pm$ 0.21 & 93.88 $\pm$ 0.09 &
			77.36 $\pm$ 0.21 & 89.14 $\pm$ 0.14 &
			90.42 $\pm$ 0.15 & 97.05 $\pm$ 0.07 \\
			ARF-SFR-Net$_{\rho_{\max}=5}$ & ResNet-12 &
			84.97 $\pm$ 0.18 & 94.21 $\pm$ 0.10 &
			77.88 $\pm$ 0.26 & 89.53 $\pm$ 0.11 &
			90.86 $\pm$ 0.14 & 97.31 $\pm$ 0.05 \\
			ARF-SFR-Net$_{\rho_{\max}=7}$ & ResNet-12 &
			85.54 $\pm$ 0.19 & 94.56 $\pm$ 0.08 &
			\textbf{ 78.68 $\pm$ 0.22} & 89.87 $\pm$ 0.17 &
			91.03 $\pm$ 0.13 & 97.48 $\pm$ 0.08 \\
			ARF-SFR-Net$_{\rho_{\max}=9}$ & ResNet-12 &
		    86.08 $\pm$ 0.17 & \textbf{94.83 $\pm$ 0.08} &
			78.65 $\pm$ 0.20 & 90.20 $\pm$ 0.10 &
			\textbf{91.24 $\pm$ 0.13} & \textbf{97.62 $\pm$ 0.05} \\
           ARF-SFR-Net$_{\rho_{\max}=11}$ & ResNet-12 &
			\textbf{86.21 $\pm$ 0.23} & 94.65 $\pm$ 0.16 &
			78.60 $\pm$ 0.19 & \textbf{90.31 $\pm$ 0.12} &
			91.02 $\pm$ 0.15 & 97.53 $\pm$ 0.06 \\
			\bottomrule
		\end{tabular}
	}
	\label{t6}
\end{table}

\subsection{The Sizes of the Receptive Fields of the Designed ARF}
After training, the sizes of the receptive fields of the designed ARF strategy on the Conv-4 and ResNet-12 backbones on the CUB-200-2011, Stanford Dogs, and Stanford Cars datasets are summarized in Table~\ref{t}. It can be seen from Table~\ref{t} that there exist significant differences in the receptive field sizes corresponding to the spatial and frequency domain branches after training. Furthermore, the receptive field sizes also differ for different FSFGIC tasks.

\begin{table}[htbp] 
	\centering 
	\caption{The size of receptive field of different branch after training on the proposed ARF-SFR-Net with Conv-4 and ResNet-12 using the CUB-200-2011, Stanford Dogs, and Stanford Cars datasets for 5-way tasks.} 
	\setlength{\tabcolsep}{4pt} 
	\renewcommand{\arraystretch}{1.} 
	\scriptsize
	\begin{tabular}{l c c c c} 
		\hline 
		Datasets & Model & Backbone & Layer1 & Layer2 \\ 
		\hline 
		\multirow{2}{*}{CUB-200-2011} & ARF-SFR-Net$_S$ & Conv-4 &3$\times$5 &7$\times$3 \\ & ARF-SFR-Net$_F$ & Conv-4 & 3$\times$5 &3$\times$5 \\ 
		\hline 
		\multirow{2}{*}{Stanford Dogs} & ARF-SFR-Net$_S$ & Conv-4 &5$\times$5 &3$\times$5 \\ & ARF-SFR-Net$_F$ & Conv-4 &5$\times$5 &5$\times$5 \\ \hline \multirow{2}{*}{Stanford Cars} & ARF-SFR-Net$\_{S}$ & Conv-4 &5$\times$3 &5$\times$5 \\ & ARF-SFR-Net$\_{F}$ & Conv-4 &5$\times$3 &3$\times$5 \\ 
		\hline 
		\multirow{2}{*}{CUB-200-2011} & ARF-SFR-Net$_S$ & ResNet-12 &3$\times$3 &7$\times$7 \\ & ARF-SFR-Net$_F$ & ResNet-12 &5$\times$3 &3$\times$3 \\ 
		\hline 
		\multirow{2}{*}{Stanford Dogs} & ARF-SFR-Net$_S$ & ResNet-12 &7$\times$5 &3$\times$5 \\ & ARF-SFR-Net$_F$ & ResNet-12 &7$\times$7 &7$\times$3 \\ 
		\hline 
		\multirow{2}{*}{Stanford Cars} & ARF-SFR-Net$_S$ & ResNet-12 &5$\times$3 &5$\times$5 \\ & ARF-SFR-Net$_F$ & ResNet-12 &5$\times$9 &3$\times$9 \\ 
		\bottomrule 
	\end{tabular} 
	\label{t} 
\end{table}

\subsection{Grad-CAM Visualization}
\begin{figure*}[!htbp]
	\centering
	\includegraphics[width=1\linewidth]{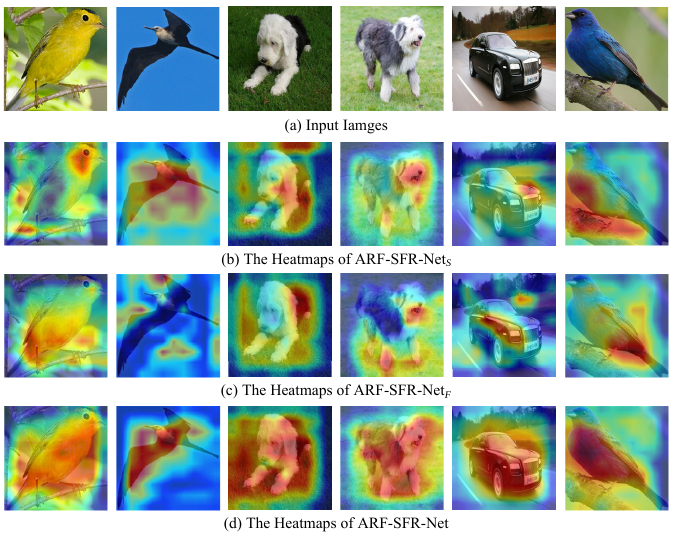}
	\caption{The heatmaps of six images visualized by different domain.}
	\label{fig5}
\end{figure*}

To further demonstrate the effectiveness of the proposed ARF-SFR-Net, we visualize model attention using Grad-CAM~\cite{selvaraju2017grad} with ResNet-12. Six sample images from Fig.~\ref{fig5}(a) are used for illustration, where higher energy regions in the attention maps correspond to more discriminative image parts. Fig.~\ref{fig5}(b)–(d) show attention maps for three variants: ARF-SFR-Net$_S$ (spatial only), ARF-SFR-Net$_F$ (frequency only), and the full ARF-SFR-Net (spatial-frequency). The results indicate that the spatial-frequency version focuses more precisely on the classification targets compared to the single-domain variants.

\section{Conclusion}
This paper proposes ARF-SFR-Net, a novel Adaptive Receptive Field-based Spatial-Frequency Reconstruction Network for few-shot fine-grained image classification. The model dynamically adjusts receptive fields in both spatial and frequency domains to generate discriminative descriptors. It then fuses these features adaptively to enable robust reconstruction-based matching. By predicting convolutional parameters, the adaptive receptive field mechanism captures flexible multi-scale patterns, enhancing feature robustness. Extensive experiments on five benchmarks demonstrate that ARF-SFR-Net consistently outperforms state-of-the-art methods, confirming its effectiveness for data-scarce fine-grained recognition.




\bibliographystyle{elsarticle-num}
\bibliography{Ref}

\end{document}